\documentclass{article}
\usepackage{ijcai19}
\usepackage{multirow}

\usepackage{times}
\usepackage{soul}
\usepackage{url}
\usepackage[hidelinks]{hyperref}
\usepackage[utf8]{inputenc}
\usepackage[small]{caption}
\usepackage{graphicx}
\usepackage{amsmath}
\usepackage{booktabs}
\usepackage{algorithm}
\usepackage{algorithmic}
\title{Predictive Ensemble Learning with Application to Scene Text Detection}

\author{
Danlu Chen$^1$\and
Xu-Yao Zhang$^2$\and
Wei Zhang$^3$\and
Yao Lu$^1$\and
Xiuli Li$^4$\and
Tao Mei$^3$
\affiliations
$^1$Beijing Laboratory of Intelligent Information Technology, Beijing Institute of Technology\\
$^2$NLPR, Institute of Automation, Chinese Academy of Sciences\\
$^3$AI Research of JD.com\\
$^4$Deepwise AI Lab, Peng Cheng Laboratory
}

\begin{document}

\maketitle
\vspace{-0.2cm}

\begin{abstract}
 Deep learning based approaches have achieved significant progresses in different tasks like classification, detection, segmentation, and so on. Ensemble learning is widely known to further improve performance by combining multiple complementary models. It is easy to apply ensemble learning for classification tasks, for example, based on averaging, voting, or other methods. However, for other tasks (like object detection) where the outputs are varying in quantity and unable to be simply compared, the ensemble of multiple models become difficult. In this paper, we propose a new method called Predictive Ensemble Learning (PEL), based on powerful predictive ability of deep neural networks, to directly predict the best performing model among a pool of base models for each test example, thus transforming ensemble learning to a traditional classification task. Taking scene text detection as the application, where no suitable ensemble learning strategy exists, PEL can significantly improve the performance, compared to either individual state-of-the-art models, or the fusion of multiple models by non-maximum suppression. Experimental results show the possibility and potential of PEL in predicting different models’ performance based only on a query example, which can be extended for ensemble learning in many other complex tasks.
\end{abstract}

\section{Introduction}
Nowadays, deep learning has became the dominant solution for many different tasks, from the \emph{traditional pattern recognition} (or classification) task to more \emph{complicated tasks} like object detection, image segmentation, speech recognition, machine translation, visual question answering, and so on. In classification task, the output is usually a fixed-length probabilities representing the belongings to different classes, and therefore, combining the predictions from multiple models is easy and straightforward. However, in other tasks where multiple outputs are not easily-comparable (like bounding boxes) or even not well-aligned (like variable-length sequences in speech or language tasks), the combination becomes difficult, and this is the reason why there is rare research on multi-model combination in these complex tasks.

Ensemble learning is a widely-studied topic in literature, and most researches are focused on classification tasks, with many famous strategies like bagging~\cite{Breiman1996bagging}, boosting~\cite{Freund1997boosting}, error-correcting output coding~\cite{Dietterich1994ECOC}, and so on. Moreover, there are many efficient and effective approaches to combine classification posterior probabilities for ensemble-based decision-making~\cite{Xu1992combining_classifiers}: from the simple averaging and majority voting based approaches, to more advanced methods like belief integration based on Bayesian formula, decision fusion with Dempster-Shafer theory, and so on.

However, for other tasks which are more complicated than classification, it is hard to apply ensemble learning directly. For example, in speech and language domains like speech recognition and machine translation, different models will output different textual sequences with variable lengths, therefore, it is hard to combine them. To deal with this, one solution is the combination at frame-level~\cite{Chebotar2016speech_ensemble} by taking a weighted average of different models' state posteriors, which requires time-synchronous predictions for different models. To make ensemble at final-result level, like combining ``she went school'' and ``she was at school'', more complicated strategies should be considered, like the sentence-level, phrase-level, word-level combinations~\cite{Rosti2007translation_combination}, where the alignment of different hypotheses~\cite{Ayan2008alignment_translation} is important. Moreover, joint evaluation of several models is usually time-consuming, and distilling multiple models into a single student model is a widely-used strategy~\cite{Chebotar2016speech_ensemble}.

\begin{figure*}[t!]
    \centering
    \includegraphics[width=0.99\textwidth]{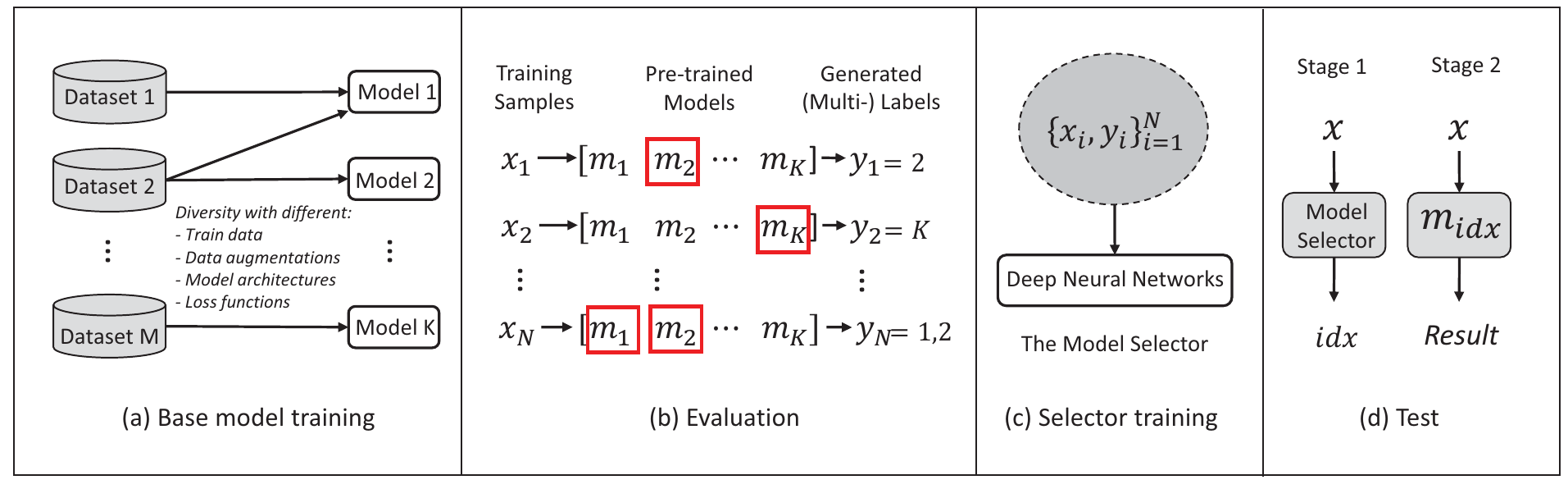}
    \caption{An illustration of predictive ensemble learning (PEL).}
    \label{fig:framework}
\end{figure*}

In another task of object detection which outputs a lot of bounding boxes for an input image, the ensemble learning is also difficult. First, it is usual that different models will predict different number of bounding boxes, and second, how to effectively compare and merge multiple boxes is also not clear. A simple solution is directly putting all bounding boxes from different models together, and then applying non-maximum suppression (NMS)~\cite{Neubeck2006nms} to reduce redundancy. However, the number of false detection may increase significantly and result in unexpected downhill performance. Since it is hard to find effective ensemble strategies for detection, many researches have made attempts from other perspectives. The detector ensemble~\cite{Dai2007part_detector} seeks a hierarchical group of part and substructure detectors to build final detection. The rank of experts~\cite{Bae2017rank_experts} selects detectors with high precision among trained detectors for each class to build class-wise ensemble. \cite{Malisiewicz2011ensemble_svm} applies ensemble of exemplar-SVMs for object detection. \cite{Wei2018fusion_ensemble} proposes fusing multiple detection results using measure of agreement and axis-aligned bounding box fuzzy integral. However, a simple and effective method for ensemble of detection models is still lacking.

In this paper, we propose a new approach named predictive ensemble learning (PEL) to handle these difficulties. The basic motivation is simple and straightforward:
\begin{quote}
\emph{Using powerful prediction ability of deep neural networks as the selector, to predict the index of the best performing model, among a pool of base models, based only on the query example itself, and thus transforming ensemble learning to a traditional classification task.}
\end{quote}
The rationale of PEL is that: although models in the pool are complementary, not every base model can serve as expert in dealing with all samples, and hence PEL is to measure the matching degree between sample and model, and then let each base model perform its own adept job. The advantages of PEL are obvious. First, it avoids the difficulties on combining complex outputs as described above, since example-specific selection of a single best model is performed, and only the result of the selected model is used. Moreover, it also avoids joint evaluation of all models, because only two models (the selector plus the selected model) need to be tested for each evaluation, which will significantly improve computational speed. Ideally, PEL is a general method and can be used for different tasks with task-specific criteria on defining the best performing model (like precision, recall, F-measure, BLUE, and so on). The only concern is actually on the accuracy of the selector in PEL, which has high impacts on the final performance. Since deep neural network performs really good at classification, this is hoped to work well.

A similar idea previously studied in literature is dynamic classifier selection (DCS)~\cite{woods1997combination}, by estimating each classifier's accuracy in local regions of feature space surrounding an unknown test sample and then using the decision of the most locally accurate classifier. The classifiers are selected on the fly, according to each query sample to be classified. DCS works by accessing the competence level of each base classifier, and only the most competent one would be selected to predict the label of a specific new sample. Researches on DCS have been conducted from both theory~\cite{kuncheva2002theoretical} and practice~\cite{fernandez2014we}. However, most of them are focused on traditional methods (with feature extraction and classifier design) and only applied to classification ensemble. In this paper, we focus on end-to-end deep neural networks which should work better, and consider more complicated ensemble tasks.

To evaluate the effectiveness of PEL, we take scene text detection as an application case, because no suitable ensemble learning method exists, but meanwhile, a branch of diverse models have been proposed in this field, like CTPN~\cite{tian2016detecting}, EAST~\cite{zhou2017east}, PixelLink~\cite{deng2018pixellink}, Mask TextSpotter \cite{lyu2018mask}, and so on. Extracting text from scene image provides additional significant information, benefiting a wide range of applications. However, scene texts usually have complex backgrounds and appear with various shapes, sizes, fonts, colors, aspect ratios, orientations, and so on. It is quite difficult to build a single expert model for covering all these situations. Particularly, some approaches may master detecting small compact text instance, while some methods do well in long text-line, and other works are more suitable for detecting irregular text regions. Therefore, PEL can gather the advantages of each method, enhancing the overall performance of scene text detection. Experimental results on several benchmarks of ICDAR 2013, ICDAR 2015 and SCUT-FORU have shown the effectiveness and potential of PEL.

\section{Predictive Ensemble Learning}\label{sec:pel}
In this section, we describe the general framework of predictive ensemble learning (PEL), which is illustrated in Fig.~\ref{fig:framework}, and includes four steps as follows.

\textbf{Base Model Training.} In deep learning era, it is frequent to see different models with incremental improvements being gradually proposed for a particular task. Normally, the latest proposed model usually reports highest performance. However, it is unwise to ignore previously proposed models, because different models are usually complementary to each other, due to their diversities as shown in Fig.~\ref{fig:framework}a. The first and most important diversity is coming from the network architectures and training objectives, which are usually viewed as the main contribution when evaluating a new work. The randomness is widely-used in deep learning to improve generalization performance such as dropout, data augmentation, and so on. This will also increase the diversity among models. At last, even for the same task, there are always multiple released datasets collected with different conditions, and training with different datasets provides the most straightforward diversity among models.

\textbf{Evaluation.} After obtaining the base models, the second step is the evaluation of their performance. As shown in Fig.~\ref{fig:framework}b, for each training sample, all the models are ranked according to a particular task-specific measurement or scoring function, and then the index of the best-performed (top ranked) model is chosen as the label for this sample. Since this is a sample-level model selection, it is often that several models getting the same score for a particular sample. In this case, the indexes of multiple models are assigned simultaneously to this sample, resulting in a multi-label problem. After evaluation, we actually get a new classification dataset, which is then used to train the selector.

\textbf{Selector Training.} Actually, there are multiple choices for defining the selector, for example, using a learning-to-rank model to predict the orders of different models. However, since classification task is easier to train and deep neural networks are good experts at classification, we define the selector as a deep neural network based classifier as shown in Fig.~\ref{fig:framework}c. The training of the selector depends on previous collected dataset in Fig.~\ref{fig:framework}b, for example, softmax-criteria can be used for single-label dataset, while the one-vs-all training and sigmoidal activation are more suitable for multi-label prediction.

\begin{figure}[t!]
    \centering
    \includegraphics[width=0.85\columnwidth]{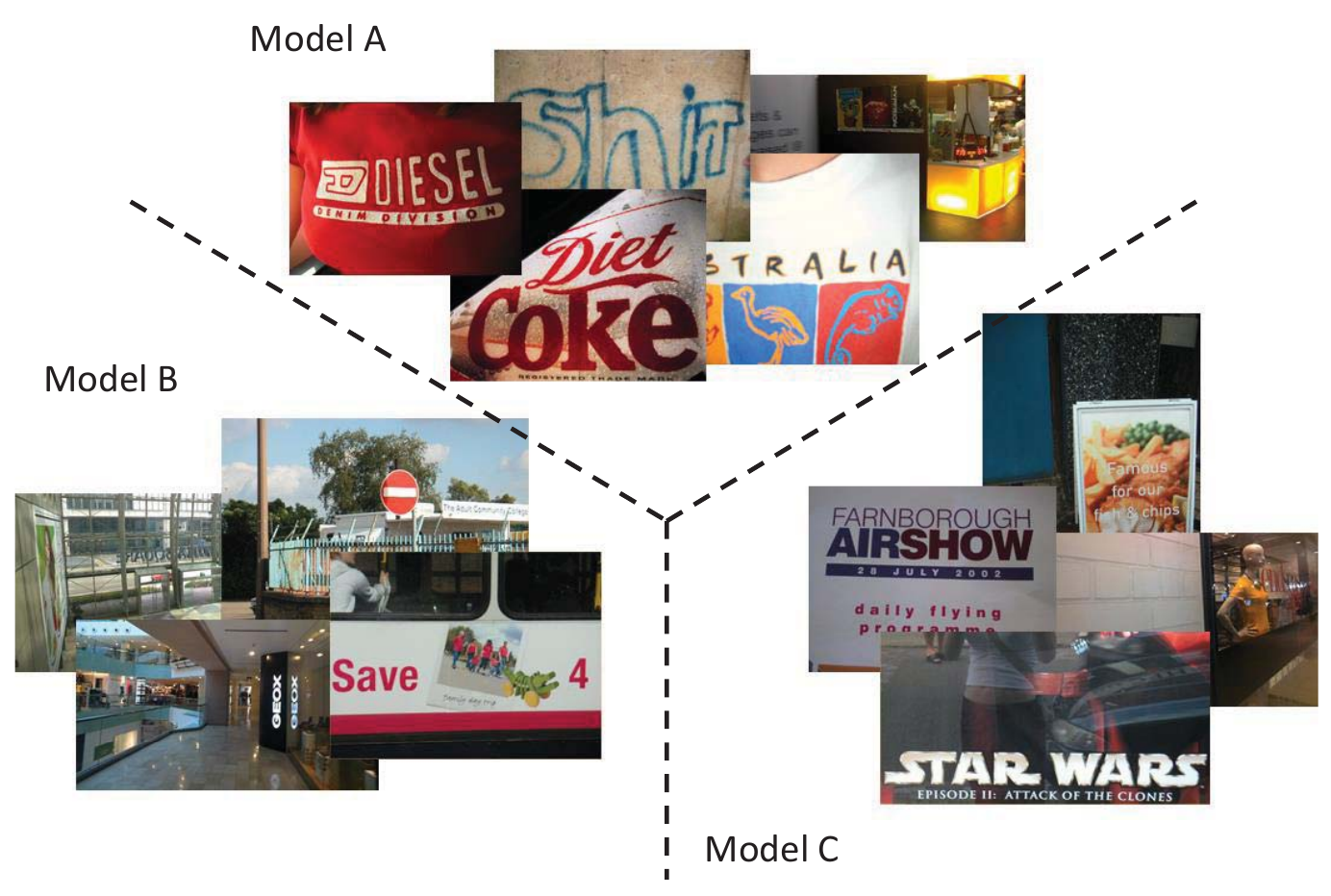}
    \caption{The model selection task for scene text detection.}
    \label{fig:selector}
\end{figure}

Fig.~\ref{fig:selector} shows the selector in the scene text detection task, which should partition images into different model categories. The classification task here is actually not intuitive compared with other classification tasks where human are usually experts. It is hard to judge how the classification is performed and which features are critical. However, with large enough data, we can handle this using end-to-end deep neural networks. Since classification is a well-studied topic in deep learning, many strategies can be used to pursue as high precision as possible for the selector.

\textbf{Two-stage Test.} The test phase of PEL, as illustrated in Fig.~\ref{fig:framework}d, contains two stages: the model selection stage and the selected model execution stage. In the first stage, for each query sample, we always predict a single index with highest score given by the selector, no matter the selector is trained with single-label or multi-label datasets. In the second stage, the selected model is then evaluated on the same query sample to obtain final result.

Since only a single sample-specific selected model is used in testing, we need not to care about the particular forms of the model outputs, which may be difficult to combine when multiple results exist. In other words, for PEL, the task-specific models can be viewed as black-boxes. Only the protocol on evaluating model performance are required (Fig.~\ref{fig:framework}b), for transforming the ensemble problem to a classification task of selector training (Fig.~\ref{fig:framework}c). Another advantage of the two-stage test is to avoid the evaluation of all base models, which will reduce a lot of computational burdens, especially considering that deep learning models are usually time-consuming and resource-hungry.

\section{PEL for Scene Text Detection}
In this section, we apply PEL for a particular task of scene text detection, where many different models have been proposed, but no suitable ensemble learning strategy exists.

\subsection{Brief Review of Scene Text Detection Models}\label{sec:review}
In recent years, deep learning based algorithms become the mainstream in text detection. Connectionist text proposal network (CTPN)~\cite{tian2016detecting} employs a vertical anchor mechanism to predict a sequence of fine-scale text components to be connected through heuristic rules. TextBoxes~\cite{liao2017textboxes} follows the standard object detection network of SSD~\cite{liu2016ssd} and adopts anchors of large aspect ratio and irregular shape to fit oriented scene text line. EAST~\cite{zhou2017east} integrates all feature maps together by a U-Net~\cite{ronneberger2015u} structure and performs dense predictions by directly regressing the boundary offsets from a given point. Rotation region proposal network (RRPN)~\cite{ma2018arbitrary} draws inspiration from standard Faster R-CNN~\cite{ren2015faster} and adds rotation to anchors and RoIPooling. SegLink~\cite{shi2017detecting} and PixelLink~\cite{deng2018pixellink} detect text segments and then link them into complete text instances by link predictions. Corner localization methods~\cite{lyu2018multi} learns to detect the corner of text instance, and groups the corners into the text instance by their relative position. Mask TexstSpotter~\cite{lyu2018mask} proposes a modification of Mask R-CNN~\cite{he2017mask} to produce the existence and locations for each single character. Feature enhancement network (FEN)~\cite{zhang2018feature} designs the network with a fusion of different level semantic information, enhancing the text feature for region proposal.

\subsection{Evaluation Measurement}\label{sec:measurement}
A key step in PEL is the model evaluation to prepare dataset for selector training (Fig.~\ref{fig:framework}b). Multiple strategies can be used to represent a text region like bounding box, rotated rectangle, quadrangle, and even irregular shape. In text detection, due to the diversity of models, different model usually outputs different number of detected regions.

Suppose a particular model predicts $N_{det}$ text regions $\{d_1, d_2, \ldots, d_{N_{det}}\}$ for a image, while the ground truth of this image contains $N_{gt}$ text regions $\{g_1,g_2,\ldots,g_{N_{gt}}\}$. We can match every predicted text region to each ground truth with the intersection-over-union (IOU) between them:
\begin{equation}
    \text{IOU}(d_i, g_j) = \frac{|d_i \cap g_j|}{|d_i \cup g_j|},
\end{equation}
where $|\cdot|$ denotes the area of the polygon. By setting a pre-defined threshold $\tau$ (generally 0.5), we can define a specific detection result $d_i$ to be matched or not by:
\begin{eqnarray}
&\text{Matched:} &\exists j: \text{IOU}(d_i, g_j) > \tau, \\
&\text{Not matched:}  &\forall j: \text{IOU}(d_i, g_j) \leq \tau.
\end{eqnarray}
In this way, we can count how many predicted results are matched to the ground truth, and this number is denoted as $N_{match}$. Since there may exist duplicated or missed detection, two measurements should be used: the precision ($P$) and recall ($R$):
\begin{eqnarray}
&P = N_{match} / N_{det}, \\
&R = N_{match} / N_{gt}.
\end{eqnarray}
To fuse these measurements into a single one, the F-score ($F$) is widely-used~\cite{wang2011end} as the final measurement:
\begin{equation}
    F = \frac{2\times P \times R}{P + R}.
\end{equation}
For each training sample $x$, there are $K$ models resulting in $K$ scores $\{F^1, F^2,\ldots, F^K\}$. We use a label vector $y=\{y^1, y^2, \ldots, y^K\}$ to denote a model should be selected or not:
\begin{equation}
    y^i =
    \begin{cases}
    1, & F^i = \max_{j=1}^K F^j \quad \text{and} \ F^i \neq 0, \\
    0, & \text{Otherwise}.
    \end{cases}
\end{equation}
Only the models who get the highest evaluation score will be set as positive. When multiple models having the same highest F-score on a particular sample, we will assign all these models as positive, resulting in multiple labels for this sample. Moreover, in some particular conditions, a sample may have no best model, for example $F^i=0, \forall i$ which means none of those base models can handle well on this sample. In this case, all labels are set to zero $y^i=0, \forall i$ denoting that this is a negative example for all the models.

\subsection{Selector Network}
Another important step in PEL is the training of the selector (Fig.~\ref{fig:framework}c). When using deep learning for computer vision, VGGNet is widely chosen as backbone, we hence apply the VGG16 pre-trained on ImageNet as the feature extractor, and then add some particular fully-connected layers specialized for our task. The whole network is then re-trained on our model selection dataset prepared in above section.

Since a multi-label classification task is considered, we use \emph{sigmoid} (other than softmax) activation function in the last output layer. The number of neuron in last layer is equal to the number of base detectors $K$. Therefore, for each input image, the network will output $K$ scores between 0 and 1, representing the possibility that a model should be selected or not. To efficiently and effectively train this network, we adopt a \emph{class-balanced binary cross-entropy loss}~\cite{xie2015holistically}. The total loss can be formulated as the summation of all category-specific losses: $L = \sum_{i=1}^{K} L_i$ where $L_i$ is category-specific cross-entropy loss:
\begin{equation}
    L_i = -\beta Y^*\log{Y}-(1-\beta)(1-Y^*)\log(1-Y),
\end{equation}
and $Y^* \in \{0,1\}$ is the ground truth and $Y \in [0,1]$ is the prediction of VGG16, for a specific output neuron. For some base models, there may be a huge gap between the number of positive and negative samples. Therefore, the parameter $\beta$ named positive weight \cite{zhou2017east} is essential to balance the positive and negative samples, given by:
\begin{equation}
\beta = 1 - \frac{\sum_{y \in Y^*}y}{|Y^*|}.
\end{equation}

Actually, this can be viewed from the one-vs-all training perspective, where multiple binary classification tasks (a sample is positive or negative for a model) are trained for each model to select samples. However, these binary tasks are not independent because the same feature representation network are shared between them. This is a simple and effective approach for multi-label training. In the test phase, to avoid the difficulty on combining models, only a single model is predicted from the selector, whose details and advantages have been clarified in Section~\ref{sec:pel}.

\begin{table*}[t!]
    \centering
    \begin{tabular}{|l|c|c|c|c|c|c|c|c|c|}
        \hline
        \multirow{2}{*}{Method}                     & \multicolumn{3}{c|}{ICDAR 2013}               & \multicolumn{3}{c|}{ICDAR 2015}               & \multicolumn{3}{c|}{SCUT-FORU}                \\
        \cline{2-10}
                                                    & Precision     & Recall        & F-score       & Precision     & Recall        & F-score       & Precision     & Recall        & F-score       \\
        \hline
        SSTD~\shortcite{he2017single}               & 89            & 86            & 88            & 80            & 73            & 77            & 75.6          & 90.9          & 82.5          \\
        PixelLink~\shortcite{deng2018pixellink}     & 86.4          & 83.6          & 84.5          & 85.5          & 82.0          & 83.7          & 53.9          & 77.1          & 63.4          \\
        Mask TextSpotter~\shortcite{lyu2018mask}    & 94.1          & 88.1          & 91.0          & 85.8          & 81.2          & 83.4          & 59.7          & \textbf{96.0} & 73.6          \\
        NMS                                         & 76.7          & \textbf{89.7} & 82.7          & 57.9          & 82.3          & 68.0          & 50.0          & 94.2          & 65.3          \\
        \hline
        Proposed Method                             & \textbf{94.5} & 89.6          & \textbf{92.0} & \textbf{86.5} & \textbf{84.5} & \textbf{85.5} & \textbf{75.7} & 92.4          & \textbf{83.2} \\
        \hline
    \end{tabular}
    \caption{Comparisons with base detectors on ICDAR 2013, ICDAR 2015 and SCUT-FORU.}
    \label{tab:ensemble}
\end{table*}

\section{Experiments}

\subsection{Base Detectors}
There are many methods in text detection as described in Section~\ref{sec:review}. In our experiments, we choose three state-of-the-art methods for PEL, due to their good performance and also being quite unlike from each other. \textbf{Single Shot Text Detector (SSTD)}~\cite{he2017single} presents a text anchor-based detector methods that outputs word-level results in single-shot with a text attention module and a hierarchical inception module, which works well on multi-orientation text detection. \textbf{PixelLink}~\cite{deng2018pixellink} is based on the idea of instance segmentation by pixel-wise link prediction to learn the relationship between two adjacent pixels, which can achieve state-of-the-art results with less data and fewer iterations compared with regression-based methods. \textbf{Mask TextSpotter}~\cite{lyu2018mask} modifies the Mask R-CNN framework to acquire precise text detection and recognition via semantic segmentation, which gets impressive performance in handling text instances of arbitrary shape such as curved text.

The reason on choosing these three models is two-fold. First, they are driven from different perspectives on text detection and therefore contain great diversities among them. Second, their performance on text detection are representative and also comparable. Adding more models in PEL will increase the difficultly of selector training, due to the unbalance problem caused by some low-performance models that almost never been labeled as positive, and also the confusion and redundancy from similar models.

\subsection{Benchmark Datasets}
Three benchmark datasets are used in our experiments:

\paragraph{ICDAR 2013:} Focused scene text dataset in challenge 2 of the ICDAR 2013 robust reading competition, composed 229 training images and 233 testing images in different resolutions. The text instances are horizontal or nearly horizontal, annotated as rectangle in word.

\paragraph{ICDAR 2015:} Incidental text dataset in challenge 4 of the ICDAR 2015 robust reading competition, including 1500 pictures (1000 training and 500 test) captured by Google glasses with relatively low resolutions and multi-oriented texts. All text regions are annotated with word-level quadrangles by 8 coordinates of four clock-wise corners.

\paragraph{SCUT-FORU:} OCR universal dataset collected from Flickr and released by South China University of Technology, containing two parts: English2k and Chinese2k. We employ the English2k word-level detection subset composed of 1200 images for training and 515 for testing, respectively. Only localization annotation of words are used in our training stage.

\begin{table}[t!]
    \centering
    \begin{tabular}{|l|c|c|c|}
        \hline
        Method                                      & Precision     & Recall        & F-score       \\
        \hline
        CTPN~\shortcite{tian2016detecting}          & 93            & 83            & 88            \\
        SSTD~\shortcite{he2017single}               & 89            & 86            & 88            \\
        EAST~\shortcite{zhou2017east}               & 92.6          & 82.7          & 87.4          \\
        SegLink~\shortcite{shi2017detecting}        & 87.7          & 83.0          & 85.3          \\
        TextBox++~\shortcite{liao2018textboxes++}   & 89            & 83            & 86            \\
        PixelLink~\shortcite{deng2018pixellink}     & 86.4          & 83.6          & 84.5          \\
        Mask TextSpotter~\shortcite{lyu2018mask}    & 84.1          & 88.1          & 91.0          \\
        FEN~\shortcite{zhang2018feature}            & 93.9          & \textbf{89.7} & 91.8          \\
        \hline
        Proposed Method                             & \textbf{94.5} & 89.6          & \textbf{92.0} \\
        \hline
    \end{tabular}
    \caption{State-of-the-art results on ICDAR 2013. All methods are evaluated under the ``DetEval'' protocol with a single scale.}
    \label{tab:ic13}
\end{table}

\subsection{Implementation Details}
The optimization used in experiments is mini-batch stochastic gradient descent (SGD) by back-propagation with a momentum of 0.9 and a weight decay of 0.0005. The input image used for selector training is reshaped with size $512\times512$. The mini-batch size is 16. The initial learning rate in SGD is $10^{-3}$, it is decayed by a weight of 0.1 for every 20 epochs, and finally stops at $10^{-6}$.

Data augmentation is used in our training. First, training images are rotated in range [-15, 15] degrees randomly. And then, subimages are randomly cropped from the rotated images. During cropping, we ensure none of the text regions would be truncated. The ratio of crop is ranging from 0.1 to 1, and the aspect ratios range from 0.5 to 2. Some other augmentation tricks, such as randomly modifying the brightness, contrast, are also used.

Randomly masking, an online data augmentation strategy, is also adopted in the training stage of our method, in order to learn the detail information from each text region in scene text images. There may be several text regions in a training sample, but only part of them play a decisive role in deciding whether the base detectors are suitable or not. The text regions in a training sample are randomly labeled as “do not care” with a probability of $p_{mask}$, so it would be masked in the input image, and $p_{mask}$ decreases as training goes on.

At last, we adopt online hard example mining to better distinguish some hard patterns, by sorting the losses in a mini-batch and selecting the highest $1/2$ of batch size samples.

\begin{table}[t!]
    \centering
    \begin{tabular}{|l|c|c|c|}
        \hline
        Method                                      & Precision     & Recall        & F-score       \\
        \hline
        CTPN~\shortcite{tian2016detecting}          & 74            & 52            & 61            \\
        SSTD~\shortcite{he2017single}               & 80            & 73            & 77            \\
        EAST~\shortcite{zhou2017east}               & 83.6          & 73.5          & 78.2          \\
        SegLink~\shortcite{shi2017detecting}        & 73.1          & 76.8          & 75.0          \\
        PixelLink~\shortcite{deng2018pixellink}     & 85.5          & 82.0          & 83.7          \\
        Lyu \emph{et al.}~\shortcite{lyu2018multi}  & \textbf{94.1} & 70.7          & 80.7          \\
        Mask TextSpotter~\shortcite{lyu2018mask}    & 85.8          & 81.2          & 83.4          \\
        TextSnake~\shortcite{long2018textsnake}     & 84.9          & 80.4          & 82.6          \\
        \hline
        Proposed Method                             & 86.5 & \textbf{84.5} & \textbf{85.5} \\
        \hline
    \end{tabular}
    \caption{State-of-the-art results on ICDAR 2015.}
    \label{tab:ic15}
\end{table}

\begin{figure*}[t!]
    \centering
    \includegraphics[width=0.95\textwidth, height=0.6\textwidth]{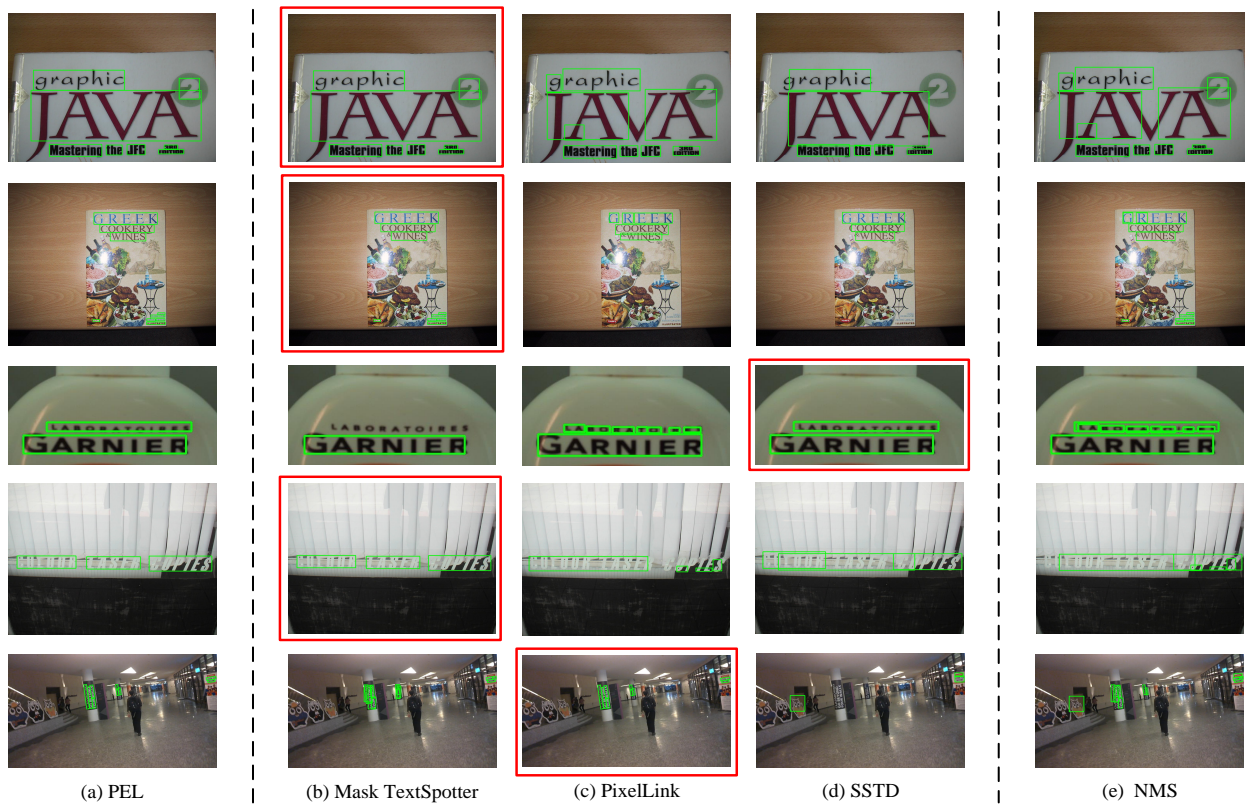}\\
    \caption{The comparison of PEL and NMS. The selected model in PEL is framed with red rectangle.}
    \label{fig:result}
\end{figure*}

\subsection{Comparison with Base Detectors}
By predicting and then executing a sample-specific best-performed model, the performance of PEL should be better than each individual base models, because each model only has its own specific skilled scenario, while PEL can dynamically select from and combine them. The comparisons of PEL with the base detectors on three benchmark datasets are shown in Table~\ref{tab:ensemble}. It is shown that the proposed method consistently outperforms each single model under all considered measurements, indicating the effectiveness of PEL in combining multiple detection models.

\subsection{Comparison with State-of-the-art}
Although PEL is learned with the ensemble of three base models in our experiment, we show that it can also outperform many other state-of-the-art models which have not been selected into the pool of base models. As shown in Table~\ref{tab:ic13}, our method reaches 0.92 in F-score, slightly higher than the best method FEN which is not used in PEL. As for ICDAR 2015, our method achieve much better results than all state-of-the-art methods as shown in Table.~\ref{tab:ic15}.

\subsection{Comparison with NMS}
As discussed before, a straightforward method for combining multiple detection result is the non-maximum suppression (NMS), whose results are also shown in Table~\ref{tab:ensemble}. The performance of NMS is far below our method and even worse than some base detectors. We illustrate the working process of PEL and NMS in Fig.~\ref{fig:result}. By selecting a best-performed model from several models, PEL can better use the special skills of individual models to make clear prediction. Contrarily, although NMS can also fuse the results of multiple models, it is not directly connected to the final measurement (Section~\ref{sec:measurement}), and therefore, will produce redundant and false detection.

\begin{table}[t!]
    \centering
    \begin{tabular}{|c|c|c|}
        \hline
        Method      & PEL   & Oracle    \\
        \hline
        ICDAR 2013  & 92.0  & 93.8      \\
        ICDAR 2015  & 85.5  & 90.7      \\
        SCUT-FORU   & 83.2  & 84.5      \\
        \hline
        Selector (train/test)  & 96.4\% / 80.6\%  & 100\% / 100\% \\
        \hline
    \end{tabular}
    \caption{The comparison of PEL and the Oracle on three datasets.}
    \label{tab:oracle}
\end{table}

\subsection{Comparison with Oracle}
The most important issue in PEL is the classification accuracy of the selector in PEL. Normally, the higher the accuracy, the better the final ensemble performance. The accuracy of the selector is 96.4\% for training and 80.6\% for test as shown in Table~\ref{tab:oracle}. If a perfect selector (100\% accuracy) exists, i.e., the \emph{Oracle}~\cite{woods1997combination}, we can guess the upper bound on the performance of PEL. As shown in Table~\ref{tab:oracle}, we are actually far from the limitations, indicating that more efforts would be paid on the selector training of PEL, for example, using more advanced architecture and training strategies. 

\section{Conclusion and Future Work}

In this paper, we proposed and verified the Predictive Ensemble Learning (PEL) for tasks which have difficulties in applying ensemble learning by conventional techniques. The main idea is to use a deep neural network classifier for dynamically predicting the most-fitted model from a pool of models for each test example. The application of PEL on scene text detection has demonstrated its advantage and potential which outperforms both individual state-of-the-art models and the combination through NMS. The proposed PEL can well handle the fusion of models which have variable, miss-aligned, or hard-to-compare outputs, and meanwhile, PEL will greatly improve the efficiency of multi-model joint evaluation. In future, we will continue to apply PEL on other tasks like speech and language domains, by using other models like recurrent networks as the selector. Although being simple and straightforward, it is our hope that PEL will be widely applied and further studied.

{\small
\bibliographystyle{named}
\bibliography{ijcai19}}

\end{document}